\documentclass[conference]{IEEEtran}

\usepackage{cite}
\usepackage{amsmath,amssymb,amsfonts}
\usepackage{graphicx}
\usepackage{booktabs}
\usepackage{url}
\usepackage{hyperref}

\hypersetup{
  hidelinks,
  pdfauthor={Rafiqul Islam},
  pdftitle={Verifying Test-State Identity in Paired Evaluation of Vision-Language-Action Models},
}

\begin{document}

\title{Verifying Test-State Identity in Paired Evaluation of\\
Vision-Language-Action Models}

\author{
\IEEEauthorblockN{Rafiqul Islam}
\IEEEauthorblockA{Independent Researcher\\
Email: rafiqul713@gmail.com}
}

\maketitle
\onecolumn

\begin{abstract}
Simulation-based VLA evaluation often treats matching task, seed, and episode identifiers as sufficient evidence that two policies faced identical test conditions.
Under specific vectorized reset and autoreset conditions, different termination timing can change subsequent initial-state assignment, so the same episode index need not mean the same test state.
We present an evaluation-integrity framework for paired comparison---invariants I1--I5, provenance logging, optional scene fingerprints, and an automated checker---and a controlled LeRobot~0.6.1 LIBERO harness study with emulated controllers.
Four evidence levels are kept separate: (A)~mixed-horizon Sync wave-1 IDs $[20,5,22,7]$ vs $[4,5,6,7]$; (B)~homogeneous separate-job Sync control (16 cells, 0 ID/fingerprint mismatches); (C)~Async+Freeze batch-size map differences (12 I3 mismatches; bs$=$8 incomplete; mechanism unisolated); (D)~a synthetic McNemar illustration only.
Controllers were emulated and no real-VLA shared-vector experiment was completed.
We recommend checking declared/resolved init-state identity and provenance before paired statistics.
We do not claim that published VLA leaderboards are generally invalid.
\end{abstract}


\section{Introduction}
Simulation benchmarks are central to robot learning and VLA evaluation.
When two policies are compared under a paired protocol, matching task, seed, and episode identifiers is often taken to mean identical test conditions.

That assumption can fail.
If Policy~A terminates earlier than Policy~B in a vectorized simulator, different reset/autoreset schedules can assign different initial states to later rollouts that still share the same episode index.
The labels look paired; the test conditions may not be.
Paired tests require genuine pairing---shared resolved test-state identity---not merely shared episode numbers.

This paper concerns \emph{evaluation integrity} for paired VLA comparisons in simulation: check resolved test-state identity and provenance rather than assume that identifiers imply pairing.

\paragraph{Case-study term}
We use Outcome-Dependent Evaluation Sampling (ODES) for the case-study condition in which subsequent test-state assignment depends on prior termination/reset history.
The mechanism is execution-condition dependent rather than universal; we do not claim discovery.

\paragraph{Relation to prior work}
Issue~\#4152~\cite{lerobot4152} documents that LIBERO \texttt{initial\_state\_id} assignment under vectorized Sync can depend on termination and reset history, not episode index alone.
PR~\#2899~\cite{lerobot2899} discusses advancing \texttt{init\_state\_id} on reset in LeRobot's LIBERO wrapper.
SmolVLA\_RL notes~\cite{smolvla_rl_repro} report related reproduction concerns for paired evaluation.
Those sources provide mechanism reports and engineering context; they do not provide reusable paired-comparison integrity invariants, automated checking with explicit \texttt{PASS}/\texttt{FAIL}/\texttt{SKIPPED} reporting, or the bounded emulated cross-mode characterization summarized as Evidence A--D here.
Paired-statistics tooling such as robostats~\cite{robostats} joins episode records on declared scenario/\texttt{init\_state\_id} keys for McNemar-style tests.
Joining on declared keys is necessary for paired inference, but does not certify that a vectorized harness assigned those keys consistently across policies or execution settings.
Our contribution is operational: (a)~invariants I1--I5 and an automated checker; (b)~provenance logging and optional scene fingerprints; (c)~controlled characterization (emulated controllers only) under mixed Sync, synchronized-horizon separate jobs, and Async batch-size variation.

We do \emph{not} claim that simulation benchmarks are generally broken, that all fast-versus-slow comparisons are mispaired, or that existing VLA papers/leaderboards are invalid.

\paragraph{Contributions}
\begin{enumerate}
\item Evaluation-integrity framework I1--I5 (I1--I3 stressed in particular harness experiments; I4--I5 enforced as checker/protocol requirements).
\item Automated checker with provenance and optional scene fingerprints.
\item Controlled LeRobot/LIBERO characterization (emulated controllers): mixed Sync divergence, synchronized-horizon separate-job control, Async batch-size map differences.
\item Practical protocol: log resolved initial-state identity and run integrity checks before paired statistics.
\end{enumerate}

\section{Problem Formulation}
\textbf{Pairing (here).}
Shared declared controls and matching identity under I1---preferring \texttt{resolved\_init\_state\_index} when both logs provide it, otherwise raw \texttt{initial\_state\_id}---optionally corroborated by a non-disabled scene fingerprint when both sides log one.
Matching episode indices alone are not sufficient.
Matching logged init IDs establish \emph{declared} identity, not necessarily identical complete simulator states; fingerprints add evidence only when their construction is documented and both values are usable.
\textbf{ODES (case-study term).}
Subsequent test-state assignment depends on prior termination/reset history~\cite{lerobot4152,smolvla_rl_repro,lerobot2899}.

\section{Evaluation-Integrity Framework}
\label{sec:framework}
We treat evaluation integrity as five checkable invariants implemented by the repository checker (\texttt{evaluation\_integrity.checker}).

\begin{itemize}
\item \textbf{I1 (Identity consistency).}
For every shared key $(\mathrm{task\_id},\mathrm{seed},\mathrm{episode\_index})$, when both rows log \texttt{resolved\_init\_state\_index} the checker compares those indices; otherwise it falls back to raw \texttt{initial\_state\_id} and states the fallback in the detail string.
It also reports \texttt{raw\_id\_mismatches} and \texttt{coverage}$=n_{\mathrm{common}}/n_{\mathrm{union\_keys}}$.
If both also log a non-empty, non-\texttt{DISABLED} \texttt{scene\_fingerprint}, those fingerprints must match.
Absent fingerprints, I1 still compares resolved (or raw) IDs but cannot corroborate scene content.
\item \textbf{I2 (Identity consistency under outcome contrast).}
If \texttt{termination\_step} or \texttt{success} differs on common keys, I1 must still hold.
If outcomes do not differ on common keys, I2 is marked \texttt{SKIPPED} (no outcome contrast; identity-under-contrast not tested).
This re-checks declared/resolved identity when outcomes differ; it is \emph{not} a causal proof of policy independence.
\item \textbf{I3 (Execution consistency).}
Caller-labeled condition logs (e.g., batch size or mode) must induce the same $(\mathrm{task},\mathrm{seed},\mathrm{episode\_index})\mapsto\texttt{initial\_state\_id}$ map relative to a reference label (\texttt{--reference LABEL}, defaulting to the first \texttt{--condition} label as given on the command line, not lexicographic order).
If fewer than two labeled conditions are supplied, I3 is marked \texttt{SKIPPED}.
\item \textbf{I4 (Execution-mode declaration).}
Each compared run must log a single non-empty \texttt{execution\_mode} (or normalized \texttt{harness\_mode}) on every row.
The checker does not validate autoreset enums beyond presence/uniqueness.
\item \textbf{I5 (Provenance completeness).}
Every row must contain non-\texttt{None} values for
\texttt{policy\_id}, \texttt{task\_id}, \texttt{episode\_index}, \texttt{initial\_state\_id}, \texttt{seed},
\texttt{batch\_size}, \texttt{slot}, \texttt{termination\_step}, \texttt{success}, \texttt{reset\_count},
\texttt{execution\_mode}, \texttt{harness\_id}.
\texttt{scene\_fingerprint} is optional and not required for I5.
\end{itemize}

\begin{table}[t]
\centering
\caption{I1--I5: role, evidence, and limits.}
\label{tab:invariants}
\begin{tabular}{@{}lp{0.28\columnwidth}p{0.28\columnwidth}@{}}
\toprule
ID & Evidence here & Limitation \\
\midrule
I1 & Sync. early/late; mixed Sync contrast & Matching IDs $\neq$ full scene proof \\
I2 & Early/late outcome contrast & I1 under contrast only; not causal independence \\
I3 & Sync/Async maps across bs & Needs $\geq$2 labeled conditions; mechanism unisolated for Async \\
I4 & Logged on audit rows & Mode presence only \\
I5 & Logged on audit rows & Protocol/checker requirement \\
\bottomrule
\end{tabular}
\end{table}

\subsection{Integrity checker}
\label{sec:checker}
The checker standardizes join keys, computes pairing fraction, localizes mismatches, checks provenance, and (when conditions are supplied) compares episode$\to$init maps across settings---capabilities that manual ID inspection does not package reproducibly.

\paragraph{Inputs}
JSON or JSONL episode logs (list, or object with an \texttt{episodes} array).
Required provenance fields are those listed under I5; optional \texttt{scene\_fingerprint}, \texttt{n\_init\_states}, and \texttt{resolved\_init\_state\_index}.
I3 needs labeled paths (\texttt{--condition LABEL=PATH}) and optional \texttt{--reference LABEL}.
Duplicate $(\mathrm{task\_id},\mathrm{seed},\mathrm{episode\_index})$ keys within a log make I1 \texttt{FAIL} (duplicated keys are listed in the detail message).

\paragraph{Outputs and exit status}
Each invariant returns exactly one status---\texttt{PASS}, \texttt{FAIL}, or \texttt{SKIPPED}---plus a detail string and evidence (I1 includes \texttt{pairing\_fraction}, \texttt{coverage}, \texttt{raw\_id\_mismatches}, and the compare mode; I3 compares labeled conditions to the reference label).
Exit code is 0 iff no invariant is \texttt{FAIL}.
\texttt{SKIPPED} means incomplete evidence for that invariant (reported as \texttt{any\_skipped}); it does not by itself make the process exit nonzero.
\texttt{fully\_checked} is true only when there is neither \texttt{FAIL} nor \texttt{SKIPPED}.

\paragraph{Example A (homogeneous Sync; I3 not exercised)}
Early vs late, \texttt{libero\_spatial} task~0, Sync bs$=$4, seed~0
(\texttt{homogeneous\_sync\_bs4.stdout.json}):
I1 pass (\texttt{pairing\_fraction=1.0}, coverage~$1.0$, $n_{\mathrm{common}}=10$; raw-ID fallback when resolved indices are absent);
I2 pass under termination contrast;
I3 \emph{skipped} (no conditions) $\Rightarrow$ \texttt{any\_skipped}=true, exit~0 if no \texttt{FAIL};
I4/I5 pass.

\paragraph{Example B (I3 exercised; cached CLI-shaped reports)}
Same-policy Async+Freeze logs with
\texttt{--condition bs1=...}, \texttt{bs2=...}, \texttt{bs4=...}
(\texttt{async\_i3\_cli.stdout.json}):
I3 \emph{fail}, 12 mismatch records vs reference \texttt{bs1} (exit nonzero).
The analogous Sync conditions (\texttt{bs1}/\texttt{bs4}/\texttt{bs8}) yield I3 \emph{pass}
(\texttt{sync\_i3\_cli.stdout.json}); I2 is \texttt{SKIPPED} in that same-policy exercise (no outcome contrast), so the run is not \texttt{fully\_checked}, but exit is~0 absent any \texttt{FAIL}.

\section{Experimental Setup}
\label{sec:setup}
\textbf{Stack.} LeRobot~0.6.1~\cite{lerobot}, Gymnasium~1.3.0, LIBERO~\cite{libero2023}; RTX~2060 6\,GB
(as recorded in \texttt{results/frozen\_phase2/environment.json}, which also records \texttt{robostats==0.1.0}).

\textbf{Modes.} Sync with \texttt{NEXT\_STEP} autoreset (finished slots reset on the next vector step); Async with \texttt{FreezeAfter} in the harness freeze path.
Batch sizes $\{1,2,4,8\}$ were attempted; Async bs~$=$8 is N/A in Table~\ref{tab:i3}.

\textbf{Controllers.} Emulated early/late (and mixed-horizon) controllers isolate harness assignment behavior; they are not learned VLAs.
\textbf{No real-VLA shared-vector experiment was completed.}

\textbf{Synchronized-horizon control (homogeneous separate jobs).}
Separate early vs late jobs on task~0 of \texttt{libero\_spatial} and \texttt{libero\_object}, seeds $\{0,1\}$, Sync batch sizes $\{1,2,4,8\}$, $n{=}10$ episodes per cell---\textbf{16 Sync cells} (2 suites $\times$ 2 seeds $\times$ 4 batch sizes), from \texttt{results/audit/summary.csv}.
In these cells all slots share the same horizon within a job, so counters advance in lockstep.

\textbf{Mixed-horizon Sync probe (Fig.~\ref{fig:odes}).}
One shared Sync vector env on \texttt{libero\_spatial} task~0, seed~0, \textbf{bs$=$4}, $n{=}8$ episodes; even slots truncate early ($T{=}8$), odd slots late ($T{=}40$).
\emph{Wave-1} denotes the second fill of the vector after the first staggered terminations/resets.

Evidence levels (not interchangeable): (A)~mixed Sync; (B)~synchronized-horizon control; (C)~Async batch-size maps; (D)~synthetic McNemar.

\section{Results}

\subsection{Evidence B: Synchronized-horizon control}
Separate early/late Sync jobs: \textbf{16 cells}, each with $n{=}10$ episodes, yield \textbf{160} early--late episode pairs.
Every pair logged a usable (non-empty, non-\texttt{DISABLED}) \texttt{scene\_fingerprint}; summary and direct log comparison report \textbf{0} fingerprint mismatches and \textbf{0} \texttt{initial\_state\_id} mismatches
(\texttt{results/audit/summary.csv}; Fig.~\ref{fig:contrast}, left).
This shows declared identity and usable fingerprints can match for homogeneous separate Sync jobs under the tested cells; it does not certify all harnesses, all homogeneous protocols, or complete simulator-state equivalence beyond the logged fingerprint definition.

\subsection{Evidence A: Mixed-horizon Sync}
\begin{figure}[t]
\centering
\includegraphics[width=0.92\linewidth]{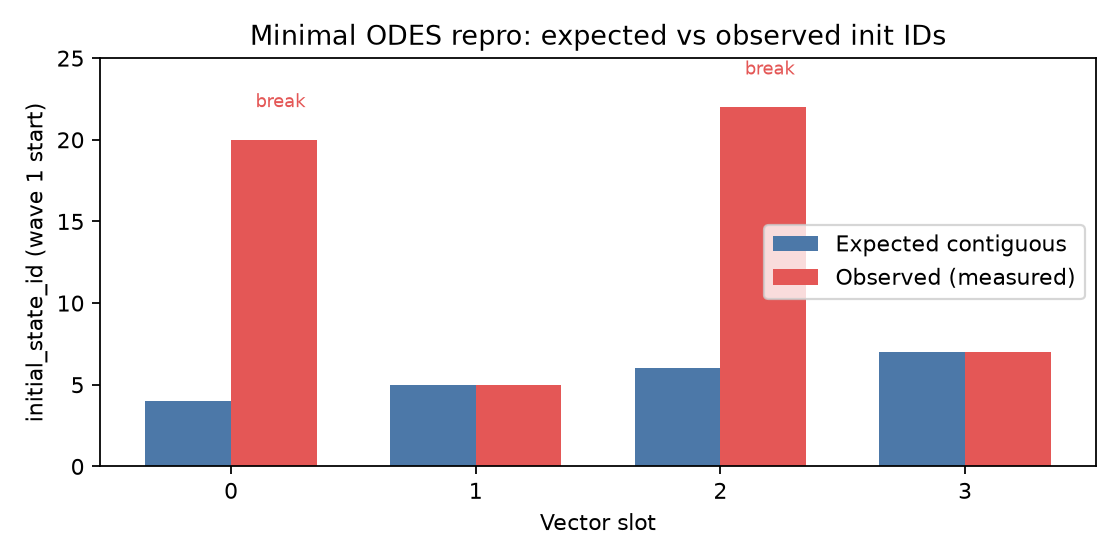}
\caption{Mixed Sync: expected vs measured wave-1 init IDs (bs$=$4).}
\label{fig:odes}
\end{figure}

\begin{figure}[t]
\centering
\includegraphics[width=0.92\linewidth]{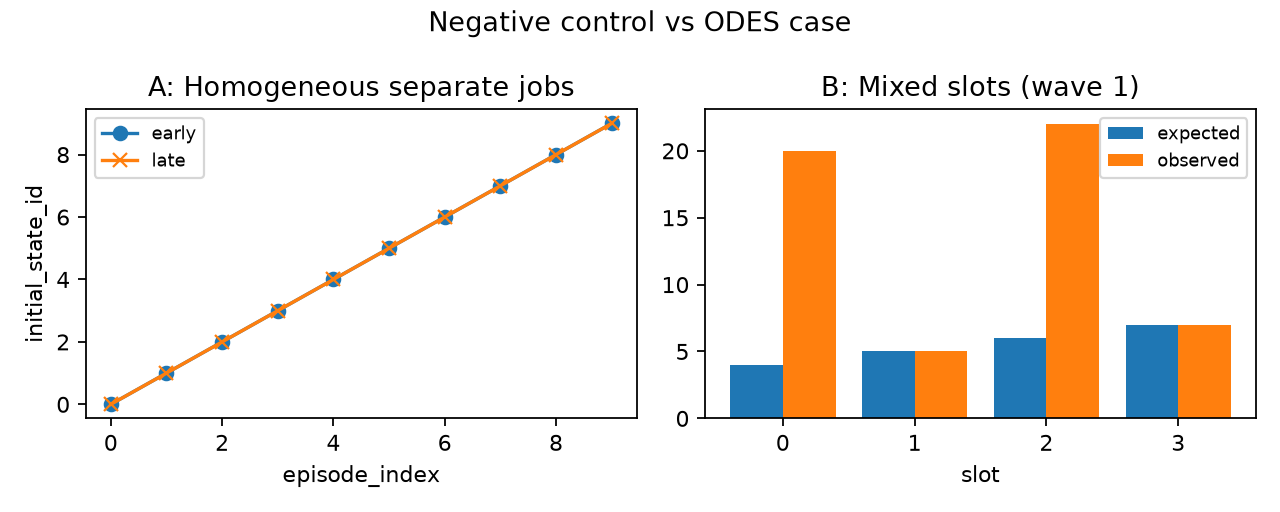}
\caption{Left: synchronized-horizon separate Sync jobs match; right: mixed Sync diverges.}
\label{fig:contrast}
\end{figure}

\textbf{Measured.} Wave-1 \texttt{initial\_state\_id} by slot: $[20,5,22,7]$ vs contiguous expectation $[4,5,6,7]$
(Fig.~\ref{fig:odes}; Fig.~\ref{fig:contrast}, right; \texttt{scripts/minimal\_repro.py} on cached grid JSON).
\textbf{Supported explanation.} Staggered truncation in one Sync+\texttt{NEXT\_STEP} vector env advances per-slot init counters unequally (consistent with public issue~\#4152 reports~\cite{lerobot4152}).
\textbf{Not established.} Prevalence across policies, suites, or non-LeRobot harnesses; learned-VLA rankings.

\subsection{Evidence C: Async batch-size sensitivity}
Table~\ref{tab:i3} asks whether, \emph{within a mode}, the episode$\to$\texttt{initial\_state\_id} map matches that mode's bs$=$1 map.
On frozen early \texttt{libero\_spatial} seed~0 logs, Sync maps for bs~$\in\{1,2,4,8\}$ are identical ($[0{..}9]$).
Async+Freeze maps differ across batch size: bs$=$1 is $[0,0,\ldots,0]$, bs$=$2 is alternating $0/1$, and bs$=$4 cycles $0..3$ (I3 fail across labeled Async conditions; 12 mismatch records in \texttt{results/checker\_examples/async\_i3\_across\_bs.json}).
Async+Freeze maps also differ from the Sync $[0{..}9]$ sequence in these logs; Table~\ref{tab:i3} does not claim Async bs$=$1 equals Sync.
Async bs$=$8 is N/A: the audit recorded the run as hung/incomplete and moved it to \texttt{results/audit/partial/} without a completed episode log (see \texttt{docs/audit\_results.md}; not a classified OOM traceback).
The mechanism that produces the Async bs~2/4 map differences was \textbf{not isolated} beyond the observed maps; whether the repeating bs$=$1 pattern originates in upstream LeRobot versus the harness FreezeAfter path is \textbf{not determined} from the frozen logs alone.

\begin{table}[t]
\centering
\caption{Within-mode: same episode$\to$init map as that mode's bs$=$1?}
\label{tab:i3}
\begin{tabular}{@{}lcc@{}}
\toprule
Mode & BS & Same as mode bs$=$1? \\
\midrule
Sync & 1--8 & yes \\
Async+Freeze & 1 & yes (reference) \\
Async+Freeze & 2 & no \\
Async+Freeze & 4 & no \\
Async+Freeze & 8 & N/A \\
\bottomrule
\end{tabular}
\end{table}

\subsection{Evidence D: Synthetic McNemar illustration}
\textbf{Synthetic illustration --- not real VLA outcomes; does not validate Evidence A--C.}
A controlled example (verified in \texttt{results/workshop/stats\_consequences\_verified.json}) shows that incorrect pairing can change paired inference:
correct init-matched pairing yields $n{=}40$, 10 discordant pairs, $\Delta{=}0.15$, exact McNemar $p{=}0.109$;
incorrect episode-index pairing yields 16 discordant pairs, $\Delta{=}0.40$, $p{=}3.05{\times}10^{-5}$ (Table~\ref{tab:stats}).
Discordant counts are constructed for illustration, not estimated from a VLA run.
This is not evidence of published false positives.
A null Monte Carlo of false-positive rate versus mismatch fraction under identical true skill is \textbf{not reported} here (no deterministic simulation artifact in the repository).

\begin{table}[t]
\centering
\caption{Synthetic illustration --- not real VLA outcomes.}
\label{tab:stats}
\setlength{\tabcolsep}{3.5pt}
\begin{tabular}{@{}lrrrrc@{}}
\toprule
Pairing & $n$ & Disc. & $\Delta$ & $p$ & @$\alpha{=}0.05$ \\
\midrule
Correct (init matched) & 40 & 10 & 0.15 & 0.109 & No \\
Incorrect (episode index) & 40 & 16 & 0.40 & $3.05{\times}10^{-5}$ & Yes \\
\bottomrule
\end{tabular}
\end{table}

\subsection{Real-policy validation}
\textbf{No real-VLA shared-vector experiment was completed.}

\section{Discussion: What Should Practitioners Change?}
(1)~Log resolved \texttt{initial\_state\_id} (and fingerprints when feasible).
(2)~Do not treat episode number alone as pairing evidence.
(3)~Treat batch size and execution mode as evaluation parameters.
(4)~Run an integrity checker before paired statistics.
(5)~Report pairing fraction and whether I3 conditions were tested.
(6)~Prefer explicit episode-indexed initial-state assignment where the harness allows it.
This is a diagnostic evaluation-methodology study for simulation benchmarks, not a real-robot transfer study.
Paired-statistics tools that join on declared init/scenario keys~\cite{robostats} remain useful; our experiment environment recorded \texttt{robostats==0.1.0}, while the citation notes a later documented API (0.2.0).
Harness-level identity checks address a complementary failure mode when those keys were assigned inconsistently.
Optional fingerprints in this harness hash init-state vector, robot state, first-observation pixels, and the raw used init id; a fingerprint mismatch therefore need not isolate pixel differences alone.

\section{Limitations}
Controllers were emulated and no real-VLA shared-vector experiment was completed, so results isolate harness behavior rather than learned-policy rankings.
Evidence is limited to \texttt{lerobot==0.6.1}, few tasks/seeds, and Async bs~8 incomplete; McNemar numbers are synthetic; Async bs~2/4 mechanisms are not isolated.
The reset/index coupling has public precedent~\cite{lerobot4152,smolvla_rl_repro}; we do not estimate prevalence across VLA benchmarks and do not claim existing papers or leaderboards are invalid.

\section{Conclusion}
VLA paired comparisons should verify declared and resolved test-state identity---and report checker outcomes including \texttt{any\_skipped} and \texttt{fully\_checked}---rather than assume that nominal episode identifiers imply genuine pairing.
Our controlled simulation case study shows that specific vectorized execution conditions can break this correspondence (Evidence A), while synchronized-horizon separate Sync jobs can preserve declared identity under the tested cells (Evidence B).
We recommend provenance logging, integrity checks, and explicit separation of observed evidence, supported explanations, synthetic illustrations, and limitations before paired statistical tests.
Reset/index coupling in LeRobot LIBERO has public precedent~\cite{lerobot4152,lerobot2899,smolvla_rl_repro}; this work does not rediscover that mechanism.

\section*{Reproducibility}
\begin{itemize}
\item \textbf{CPU reproduce (cached mixed + homogeneous cell):}
{\small\noindent\texttt{PYTHONPATH=src:.\newline
python scripts/minimal\_repro.py}\par}
Reads cached JSON under \texttt{results/audit/grid/}; runtime is seconds; no GPU.
This reproduces the wave-1 $[20,5,22,7]$ observation and one homogeneous init/fingerprint match, not the full audit grid.
\item \textbf{Checker CLI:} \texttt{PYTHONPATH=src python -m evaluation\_integrity.checker LOG\_A LOG\_B} (optional \texttt{--condition LABEL=PATH}, \texttt{--reference LABEL}). Exit code~0 iff no invariant is \texttt{FAIL}; JSON reports \texttt{any\_skipped} and \texttt{fully\_checked}.
\item \textbf{Checker unit tests:} \texttt{PYTHONPATH=src python -m pytest tests/ -q} (25 passed in this environment).
\item \textbf{Table~\ref{tab:i3} / Fig.~\ref{fig:odes} data:} frozen logs under \texttt{results/frozen\_phase2/grid\_key/}; checker reports under \texttt{results/checker\_examples/}; McNemar under \texttt{results/workshop/stats\_consequences\_verified.json}.
\item \textbf{Full harness audits:} require LeRobot~0.6.1, LIBERO, and a GPU (original runs: RTX~2060 6\,GB).
\item \textbf{Public artifact URL:} not yet available.
\end{itemize}

\bibliographystyle{IEEEtran}
\bibliography{references}

\end{document}